%% file: unigrampaper.tex
\documentclass[11pt]{article}
\usepackage[dvipsnames]{xcolor}

\usepackage{microtype}
\usepackage{graphicx}
\usepackage{subcaption}
\usepackage{booktabs} 
\usepackage{multirow}
\usepackage{times} 
\usepackage[T1]{fontenc}

\usepackage{hyperref}

\usepackage{caption}

\pdfoutput=1

\usepackage[final]{acl}

\usepackage{amsmath}
\usepackage{amssymb}
\usepackage{algorithm}
\usepackage{algpseudocodex}
\usepackage{comment}
\usepackage{calc}
\usepackage{dirtytalk} 

\usepackage[capitalize,noabbrev]{cleveref}
\newcommand{\tsp}{\textvisiblespace}

\algnewcommand{\algorithmicand}{\textbf{and}}
\algnewcommand{\algorithmicor}{\textbf{or}}
\algnewcommand{\algorithmicnot}{\textbf{not}}
\algnewcommand{\algorithmicbreak}{\textbf{break}}
\algnewcommand{\IIf}[1]{\algorithmicif\ #1\ \algorithmicthen}
\algnewcommand{\EndIIf}{\unskip\ \algorithmicend\ \algorithmicif}

\newlength{\algotitlewidth}
\newcommand{\algsubsection}[1]{%
  \Statex\makebox[\linewidth]{%
    \settowidth{\algotitlewidth}{\hspace{1ex}\textbf{#1}\hspace{1ex}}%
    \raisebox{0.35ex}{\rule{0.5\linewidth-0.5\algotitlewidth}{0.4pt}}%
    \hspace{1ex}\textbf{#1}\hspace{1ex}%
    \raisebox{0.35ex}{\rule{0.5\linewidth-0.5\algotitlewidth}{0.4pt}}%
  }
}

\newcommand{\token}{x}
\newcommand{\seq}{\mathbf{x}}
\newcommand{\vocab}{\mathcal{V}}
\newcommand{\loss}{\mathcal{L}}
\newcommand{\segs}[1]{\mathcal{S}(#1)}

\newcommand\maxlength{\ell_{\rm max}}
\newcommand\vocabsize{n}

\newcommand{\FSP}{Flat Score Pruning}

\definecolor{configurablecolor}{HTML}{2980B9} 
\definecolor{hardcodedcolor}{HTML}{C0392B}    
\newcommand{\configurable}[1]{\textcolor{configurablecolor}{{#1}}}
\newcommand{\hardcoded}[1]{\textcolor{hardcodedcolor}{#1}}
\newcommand{\mandatory}[1]{\textcolor{black}{#1}}

\newcommand{\vocabseed}{V_{\text{seed}}}
\newcommand{\vocabsizeseed}{n_{\text{seed}}}
\newcommand{\vocabsizeseedfac}{\beta_{\text{seed}}}

\newcommand{\ninterim}{n_{\text{inter}}}
\newcommand{\overshootfactor}{\alpha_{\text{inter}}}
\newcommand{\nem}{N_{\text{em}}}
\newcommand{\mprunecount}{\tau_{\text{mp}}}
\newcommand{\alphaprune}{\alpha_{\text{prune}}}

\newcommand\corpus{C}
\newcommand\corpusconcat{\corpus_{\rm concat}}
\newcommand\irank{r}
\newcommand\rcurr{\irank_{\rm curr}}
\newcommand\rstart{\irank_{\rm start}}

\newcommand\rleftboundstack{\irank_{\rm lb}}

\newcommand\suffixarray{\rm SA}
\newcommand\stack{\rm stack}

\newcommand{\goodoutlier}[1]{\textcolor{green!70!black}{#1}}
\newcommand{\badoutlier}[1]{\textcolor{red!70!black}{#1}}
\newcommand{\biggestoutlier}[1]{\underline{#1}}

\newcommand{\CC}{C\nolinebreak\hspace{-.05em}\raisebox{.25ex}{\small ++}}

\newcommand{\matchcolor}[1]{\textcolor{green!40!black}{#1}}
\newcommand{\nomatchcolor}[1]{\textcolor{red!30!black}{#1}}

\usepackage{xstring}
\newcommand{\relchange}[1]{%
  \textsuperscript{\tiny%
    \IfBeginWith{#1}{-}%
      {{\color{green!50!black}$\downarrow$\StrGobbleLeft{#1}{1}}}%
      {\IfBeginWith{#1}{+}%
        {{\color{red!70!black}$\uparrow$\StrGobbleLeft{#1}{1}}}%
        {\PackageError{relchange}{Argument must start with + or -}{}}%
      }%
  }%
}

\title{Which Pieces Does Unigram Tokenization Really Need?}
\author{
Sander Land \\
Writer, Inc. \\
\url{sander@writer.com}
   \And
   Yuval Pinter \\
   Institute for Applied AI Research \\
   Ben-Gurion University of the Negev \\
   \url{pintery@bgu.ac.il}
}

\begin{document}

\maketitle

\begin{abstract}
The Unigram tokenization algorithm offers a probabilistic alternative to the greedy heuristics of Byte-Pair Encoding.
Despite its theoretical elegance, its implementation in practice is complex, limiting its adoption to the SentencePiece package and adapters thereof.
We bridge this gap between theory and practice by providing a clear guide to implementation and parameter choices.
We also identify a simpler algorithm that accepts slightly higher training loss in exchange for improved compression.\\
\makeatletter
\ifacl@anonymize
    \raisebox{-0.15\height}{\includegraphics[width=0.3cm]{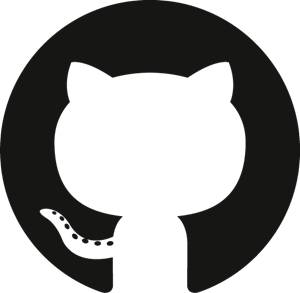}}
    \texttt{\small\,github.com/code/after/review}
\else 
    \href{https://github.com/sanderland/script_tok/}{
    \raisebox{-0.15\height}{\includegraphics[width=0.3cm]{githublogo.png}}
    \texttt{\small\,github.com/sanderland/script\_tok}
    }
\fi
\makeatother
\end{abstract}

\section{Introduction}
The Unigram language model for tokenization~\citep{kudo-2018-subword} was introduced as a principled, probabilistic alternative to the greedy heuristics of Byte-Pair Encoding~\citep[BPE;][]{sennrich-etal-2016-neural}. 
Despite its theoretical elegance, its implementation in practice is daunting: the original paper presents the training algorithm at a very high level of abstraction, online tutorials are often oversimplified to the point of being incorrect, and the only authoritative reference is a complex \CC{} implementation in SentencePiece~\citep{kudo-richardson-2018-sentencepiece}.
This has practical consequences: researchers often default to BPE or treat SentencePiece as a black box, limiting experimentation and exploration of tokenizer algorithms beyond small variations in BPE~\cite[][inter alia]{provilkov2020bpedropoutsimpleeffectivesubword,chizhov2024bpegetspickyefficient}.

This paper explains the reference Unigram training algorithm,
suggests alternatives to key steps,
and explores the effect of configurable and hard-coded parameters on tokenizer performance.
We find that costly defaults such as multiple expectation-maximization steps have little effect on the output vocabulary. 
We also present a better heuristic for constructing seed vocabularies and a simplified pruning algorithm that accepts slightly higher loss while compressing almost as well as BPE.

\section{The Unigram Model in Theory}

The Unigram model\footnote{Sometimes referred to as \emph{UnigramLM}. We use the shorter term as it is unambiguous in this context.} assumes that the probability $P(\seq)$ of a tokenized sequence $\seq = (\token_1, \dots, \token_k)$ is the product of the token probabilities $p(\token_i)$:
\begin{equation}
P(\seq) = \prod_{i=1}^k p(\token_i).
\end{equation}

When tokenizing text with a trained model, the Viterbi algorithm is typically used to determine the sequence of tokens which maximizes $P(\seq)$.
During LLM training, this can be extended to sample lower-probability token sequences for regularization. 
In tokenizer training, both the vocabulary $\vocab$ and the token probabilities $p(\token)$ for all $\token \in \vocab$ are jointly optimized to maximize the marginal log-likelihood over a training corpus $\corpus$.
For a given text $X$, let $\segs{X}$ be the set of all possible segmentations using the vocabulary.
The objective function $\loss$ to minimize is:
\begin{equation}
\loss = \frac{-1}{\sum_{X\in\corpus} |X|}\sum_{X \in \corpus} \log\left(\sum_{\seq \in \segs{X}} P(\seq) \right),
\label{eqn:loss}
\end{equation}

where $|X|$ is the length of $X$ measured in atomic tokens (e.g. bytes). We normalize by the number of atomic tokens rather than the number of texts to ensure the metric remains independent of how the corpus is subdivided. For a more detailed mathematical treatment, see \citet{clarablog}.

\section{The Unigram Model in Practice}

Since the vocabulary $\vocab$ is not known in advance, the training process must simultaneously discover a good vocabulary and estimate token probabilities.
The algorithm described by \citet{kudo-2018-subword} is an iterative procedure that starts with a large number of seed tokens and gradually reduces them to the target vocabulary size.
Both the original description and our summary of the reference implementation 
are listed in Algorithm~\ref{alg:kudo_high_level}.%

\subsection{Seed Vocabulary Initialization\label{desc:seed}}

The SentencePiece implementation uses a combination of a suffix array 
and a longest prefix algorithm to extract frequent substrings.
In short, the suffix array-based approach encodes a compact representation of the corpus' sorted suffixes, which then allows for efficient frequency counting with a stack-based emission algorithm.
\autoref{app:init} describes these steps in more detail.

Unlike many other tokenizers, SentencePiece does not pretokenize (i.e., split text into \emph{pretokens}) before training.
Instead, it operates on the entire corpus, rejecting candidate tokens during initialization based on validity criteria such as not containing internal whitespace.
As a result, the stack-based emission algorithm in the reference implementation (cf. Algorithm~\ref{alg:init} in \autoref{app:init}, lines \ref{line:stackloopstart}--\ref{line:stackloopend}) may exclude certain frequent tokens from the set of seed tokens.
This happens when the shared prefix is not a valid token, often due to trailing spaces.
For example, given the corpus \say{the\tsp{}old\tsp{}man\tsp{}the\tsp{}boat}, the substring \say{the\tsp} will be emitted, and then rejected due to its trailing space, while \say{the} is never emitted as it is not a maximal prefix.
We propose and test a \emph{maximal valid prefix recovery} variation of the algorithm, where the longest valid prefixes of rejected seed tokens are included (see Algorithm \ref{alg:init:repair} in \autoref{app:init}).
Note that in both variants, a single corpus occurrence of a word can introduce new high-frequency tokens in the seed vocabulary, by making a prefix `maximal' in at least one case.

\subsection{The EM Algorithm\label{desc:em}}

Given a fixed vocabulary, token probabilities $p(\token)$ are estimated using the expectation-maximization algorithm, 
which alternates between expectation and maximization steps.
In the {\bf E}xpectation step, the forward-backward algorithm~\cite{baum1972inequality} is used to efficiently determine the expected count of each token, averaged over all segmentations of the corpus under the current probabilities $p(\token_i)$.
The {\bf M}aximization step updates the token probabilities $p(\token_i)$ based on these expected counts. 
In SentencePiece, at the start of the M-step, tokens with an expected count below $\mprunecount=0.5$ are removed, which we refer to as the \emph{early prune step}.
This hyperparameter is sensitive to the absolute corpus size, i.e.,~duplicating a corpus would change thresholding behavior despite maintaining the same exact token distribution.

\subsection{Iterative Vocabulary Pruning}

After a fixed number $\nem$ of EM steps,\footnote{The SentencePiece implementation uses $\nem=2$.} the main pruning step is performed.
For each token, the pruning algorithm determines whether the token represents its own optimal tokenization and removes it otherwise.
For the remaining tokens, the algorithm determines each token's second-best tokenization and calculates the effect on total log-likelihood if all occurrences of the token were replaced with this second-best tokenization.
Tokens that are least critical to maintain high overall likelihood are discarded.
During this step, the probabilities $p(\token_i)$ are only used to inform the optimal tokenization paths.%
\footnote{This is a local heuristic: it cannot account for cases where, for example, \texttt{thou} is removed from the vocabulary and now a segmentation like \texttt{court|house} becomes optimal.}
The fraction of vocabulary preserved at each step is controlled by a hyperparameter $\alphaprune$ (typically set in the 0.75--0.9 range), ensuring that pruning happens gradually over several iterations. 

In the reference implementation, the second-best tokenization is determined using a generic algorithm for finding the top $k$ tokenizations. 
We note that a simple alternative is to apply the Viterbi algorithm over all paths over the token text, excluding the singleton.

\subsection{Finalization using Flat Score Pruning}

Once the iterative EM and pruning process reduce the vocabulary to at most $\overshootfactor$ times the target size (typically 1.1, i.e. 10\% over), the SentencePiece implementation performs a final vocabulary selection step.
At this stage, tokens with the lowest estimated probabilities $p(\token_i)$ are pruned regardless of token interactions.

\section{Evaluating Unigram Tokenizers}
\label{sec:evaluation}
\input{table_main_init_fsp}

In this section, we evaluate the effect of various algorithm and parameter choices on Unigram tokenizer performance.

Tokenizers are generally difficult to evaluate, as intrinsic metrics such as compression rate do not always correlate well with downstream model performance~\citep{schmidt-etal-2024-tokenization}, but are nevertheless important for inference speed.
We evaluate Unigram tokenizer training algorithms based on (1) the training objective (marginal log-likelihood, Equation \ref{eqn:loss})
and (2) compression as measured by corpus token count.
For both, lower is better.

\begin{figure*}[p] 
\centering
\begin{minipage}{\linewidth}
\begin{algorithm}[H]
\caption{Unigram Training Algorithms,
    with \hardcoded{hardcoded} and \configurable{configurable} parameters highlighted.}
\label{alg:kudo_high_level}
\input{algorithm_sp}
\end{algorithm}
\vspace{5.5pt} 
\end{minipage}

\includegraphics[width=0.97\linewidth]{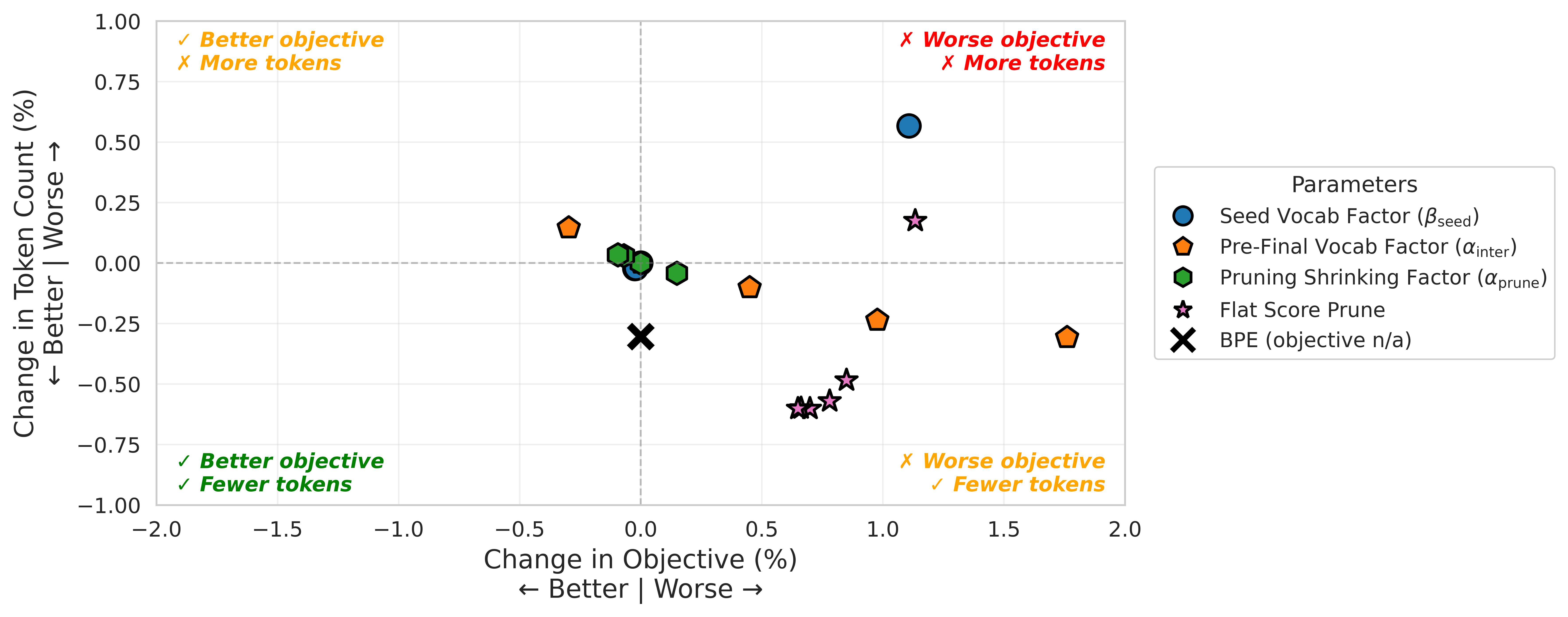}
\caption{Effect of training hyperparameters. The origin is our baseline settings, as defined in \autoref{app:hyperparams}.
Hyperparameters with minimal impact on both metrics omitted. The parameters with larger effects reveal a fundamental tradeoff: improvements in compression come at the cost of Unigram objective. \FSP{} achieves the best compression. No configuration in our search significantly improves both metrics simultaneously.}
\label{fig:scatter}
\end{figure*}

\subsection{Experimental Setup}

We test the effect of training hyperparameters using 300\,MB monolingual corpora in English, German, Korean, Chinese, Arabic, and Hindi.
Unless otherwise noted, we use SCRIPT encoding and pretokenization with character boundaries~\citep{scriptbpe} and a trained vocabulary size of 32,768 for all experiments.

\subsection{Seed Vocabulary Initialization}

We implement a pretoken-based initialization algorithm using a pretoken-to-count mapping, following the common approach in efficient BPE implementations for storing corpora.
This approach uses suffix arrays to track both text and indices, largely sidestepping the issue of incorrectly-rejected prefixes (\S\ref{desc:seed}) while improving efficiency.
We also test SentencePiece-style initialization with the \emph{maximal valid prefix recovery} variant described in \autoref{app:init}.
Results in \autoref{tab:seed_init} show that both the pretoken-count approach and the recovery variation improve both compression and objective optimization over the SentencePiece-style approach.
However, the recovery variation has only a minor effect, ranging from mildly beneficial to mildly detrimental.
We hypothesize that in larger corpora, the few remaining important seed tokens that are not caught by the corpus-based approach are likely to appear at least once, mitigating the original issue, 
while adding too many similar seed tokens may crowd out other candidates.
Therefore, in the rest of our experiments we use the more efficient pretoken-count approach without recovery.

The only other parameter in initialization is the seed vocabulary size, which we implement as a relative factor $\vocabsizeseed = \vocabsizeseedfac \cdot \vocabsize$, where $\vocabsize$ is the target vocabulary size.
While our default of $\vocabsizeseedfac=10$ is sufficient, performance drops significantly at lower values.
This is relevant given that SentencePiece's default value ($10^6$) corresponds to $\vocabsizeseedfac\approx 4$ for large multilingual tokenizers with 250k tokens.

\subsection{Iterative Vocabulary Pruning}

The main pruning loop involves several hyperparameters, many of which have minimal effect. Varying the number of EM iterations within the 1--5 range shows no consistent differences. SentencePiece applies a digamma transformation to expected counts in the M-step (cf. Algorithm~\ref{alg:kudo_high_level}) which suppresses low-count tokens; omitting this transformation yields essentially no difference. Similarly, the early prune threshold $\mprunecount$ (cf. \autoref{desc:em}) can be varied between 0 and 10 with no apparent effect.

Among the more impactful parameters, a slower shrinking factor $\alphaprune$ yields better objective values but worse compression, though both effects are small. A similar tradeoff appears when halting pruning earlier via a larger $\overshootfactor$. Motivated by these results, we investigate \FSP{} (FSP), which replaces \emph{all} pruning steps with the simpler probability-based algorithm. \autoref{fig:scatter} visualizes all parameter variations, revealing that (1) parameters that significantly affect performance exhibit a clear tradeoff with no configuration improving both metrics; and (2) FSP is the only variation which achieves compression exceeding BPE.

\autoref{tab:fsp_bpe_val} shows compression and morphological alignment results 
across vocabulary sizes (16K--64K) and in a cross-corpus setting \citep[FineWiki][1\,GB 
subsets across the same six languages]{penedo2025finewiki}. The compression--objective 
tradeoff persists across all settings. In the cross-corpus setting, BPE achieves a slight 
compression edge, which we attribute to markdown table syntax prevalent in FineWiki but 
rare in the monolingual training corpora.
We also report MorphScore~\citep[][higher is better]{arnett2025evaluatingmorphologicalalignmenttokenizers} for morphological alignment on English: Unigram methods substantially outperform BPE, consistent with Unigram's original motivation for subword regularization. Notably, FSP's compression gains come at the cost of reduced morphological alignment, positioning it between standard Unigram and BPE on this metric. \autoref{app:morphscore} presents some examples of segmentations across different tokenizers.

\section{Conclusion}

We provide a detailed guide to Unigram tokenization, clarifying implementation details that bridge the gap between the original high-level description and the SentencePiece reference implementation.

Our analysis reveals that Unigram is remarkably robust to hyperparameter choices. Several computationally expensive settings, including the number of EM iterations, digamma transformation, and early pruning thresholds, have minimal impact and can be simplified to improve training efficiency. We also demonstrate that pretoken-based initialization consistently outperforms the full-text approach in both objective and compression.

We identify a fundamental tradeoff between training objective and compression: no parameter configuration significantly improves both metrics simultaneously. Our proposed \FSP{} mechanism achieves compression closer to BPE by simplifying the pruning procedure, though at the cost of a marginally higher loss. Given the uncertain relationship between intrinsic tokenization metrics and downstream performance, this simpler alternative warrants consideration when compression is prioritized.
In cross-corpus settings, BPE achieves a slight compression edge; we observe this is driven by markdown table syntax in FineWiki, though why BPE handles this better warrants further investigation.

Overall, Unigram compression closely matches BPE, suggesting the choice between them may depend more on the need for probabilistic segmentation than on compression. We hope this work lowers the barrier to broader research and experimentation in Unigram tokenization.

\section*{Limitations}

Our evaluation focuses on intrinsic metrics (training objective and compression) rather than downstream language model performance, which remains the ultimate measure of tokenizer effectiveness. Intrinsic analysis is nonetheless the appropriate first step: as \citet{schmidt-etal-2024-tokenization} demonstrate, the relationship between tokenizer properties and downstream performance is not straightforward, making it important to first understand what the training algorithm actually does. A controlled downstream comparison would require training multiple compute-matched language models, a substantial undertaking orthogonal to our goal of clarifying and simplifying the Unigram algorithm. The most informative future experiments would evaluate FSP against standard Unigram and BPE tokenizers in matched pretraining runs, ideally across multiple languages and scales, to determine whether the compression gains translate to measurable differences in model quality or training efficiency.

\makeatletter
\ifacl@anonymize
\else 
\section*{Acknowledgments}
We thank members of the ``token \#\#ization'' Discord server for helpful discussions.
In particular, we are grateful to Robert Burke for helpful early discussions, 
Catherine Arnett for feedback and assistance with MorphScore, and Clara Meister for in-depth feedback and discussions on mathematical background. 
We also thank Karin M. Jacobsen for proofreading and helpful feedback.
Y.P. is supported by the Israel Science Foundation (grant No. 1166/23).
\fi
\makeatother

\bibliography{references}

\appendix
\onecolumn

\input{algorithm_app_init}
\clearpage
\input{tables_app_hyper}

\clearpage
\input{table_app_morphscore}

\end{document}

%% file: table_main_init_fsp.tex
\begin{table}[t]
\centering
\small
\begin{tabular}{@{}lrrr@{}}
\toprule
\textbf{Method} & \textbf{Loss} & \textbf{\#Tokens} & \textbf{Overlap} \\
\midrule
\multicolumn{4}{@{}l}{\textit{\textbf{Mean over Monolingual 30\,MB Corpora}}} \\
\addlinespace[2pt]
Pretokens (Ours) & 1.325\phantom{\relchange{+0.00}} & 5.43\phantom{\relchange{+0.00}} & ---\phantom{0} \\
Pretok., Recovery & 1.326\relchange{+0.02} & 5.43\relchange{-0.01} & 90.2 \\
Full-text (SP) & 1.333\relchange{+0.54} & 5.47\relchange{+0.75} & 79.1 \\
Full-text, Recov. & 1.326\relchange{+0.03} & 5.43\relchange{-0.01} & 88.6 \\
\midrule
\multicolumn{4}{@{}l}{\textit{\textbf{Mean over Monolingual 300\,MB Corpora}}} \\
\addlinespace[2pt]
Pretokens (Ours) & 1.337\phantom{\relchange{+0.00}} & 54.02\phantom{\relchange{+0.00}} & ---\phantom{0} \\
Pretok., Recovery & 1.338\relchange{+0.02} & 54.02\relchange{+0.00} & 91.6 \\
Full-text (SP) & 1.338\relchange{+0.05} & 54.05\relchange{+0.05} & 95.7 \\
Full-text, Recov. & 1.338\relchange{+0.03} & 54.02\relchange{-0.01} & 91.6 \\
\bottomrule
\end{tabular}
\caption{Seed vocabulary initialization comparison. Pretoken-based (Ours) vs. full-text (SentencePiece-style), with and without \emph{maximal valid prefix recovery} (Algorithm~\ref{alg:init:repair}). Superscripts show relative change (\%) vs. baseline ($\uparrow$ worse, $\downarrow$ better). Overlap: vocabulary overlap with baseline (absolute \%). Pretoken-based initialization achieves better loss and compression on both corpus sizes. The recovery variation has minimal effect, suggesting that missing prefixes are recovered naturally in sufficiently large corpora.}

\label{tab:seed_init}
\end{table}
\newcommand{\tabtwosep}{\addlinespace[4pt]\midrule\addlinespace[4pt]}
\begin{table}[t]
\centering
\small
\begin{tabular}{@{}llrrr@{}}
\toprule
\textbf{Method} & $\vocabsize$ & \textbf{Train Tok.} & \textbf{FW Tok.} & \textbf{Morph.} \\
\midrule
\addlinespace[3.3pt]
Baseline & 16K & 58.7\phantom{\relchange{+0.0}} & 65.0\phantom{\relchange{+0.0}} & 0.620 \\
         & 32K & 54.0\phantom{\relchange{+0.0}} & 59.6\phantom{\relchange{+0.0}} & 0.611 \\
         & 64K & 50.6\phantom{\relchange{+0.0}} & 55.7\phantom{\relchange{+0.0}} & 0.644 \\
\tabtwosep
FSP      & 16K & 58.2\relchange{-0.8} & 64.1\relchange{-1.4} & 0.497 \\
         & 32K & 53.7\relchange{-0.6} & 58.9\relchange{-1.2} & 0.514 \\
         & 64K & 50.3\relchange{-0.5} & 55.1\relchange{-1.1} & 0.559 \\
\tabtwosep
BPE      & 16K & 58.5\relchange{-0.4} & 64.0\relchange{-1.5} & 0.221 \\
         & 32K & 53.8\relchange{-0.3} & 58.8\relchange{-1.3} & 0.186 \\
         & 64K & 50.4\relchange{-0.4} & 54.9\relchange{-1.4} & 0.165 \\
\addlinespace[1.1pt]
\bottomrule
\end{tabular}
\caption{Compression (M tokens, lower is better, superscripts show \% change vs baseline at same $\vocabsize$) and morphological alignment across vocabulary sizes $\vocabsize$. Baseline is Unigram with our default settings, while FSP uses \FSP{}. \emph{Train Tok.}: trained and evaluated on same 300\,MB corpora. \emph{FW Tok.}: trained on FineWiki 1\,GB, evaluated on 300\,MB corpora. Note that the gap with baseline widens, while BPE gains a small edge over FSP. \emph{Morph.}: MorphScore boundary recall on English (higher is better), where Unigram dominates.}
\label{tab:fsp_bpe_val}
\end{table}

%% file: algorithm_sp.tex
\begin{algorithmic}
   \Statex {\bf Inputs and Parameters:}
   \Statex \quad $\mandatory{\corpus}$                           \Comment{Mandatory input: Training corpus}
   \Statex \quad $\mandatory{\vocabsize}$ \Comment{Mandatory parameter: Target vocabulary size}
   \Statex \quad $\configurable{\vocabsizeseed} \leftarrow 10^6$ \Comment{Initial seed vocabulary size (we use $\vocabsizeseed = \vocabsizeseedfac \cdot \vocabsize$)}
   \Statex \quad $\configurable{\nem} \leftarrow 2$ \Comment{Number of EM sub-iterations per pruning iteration}
   \Statex \quad $\configurable{\alphaprune} \leftarrow 0.75$ \Comment{Reduction factor for vocabulary size during pruning}
   \Statex \quad $\hardcoded{\overshootfactor} \leftarrow 1.1$ \Comment{Pre-final vocab size as multiple of $\vocabsize$}
   \Statex \quad $\hardcoded{\mprunecount} \leftarrow 0.5$ \Comment{Minimum expected count threshold in M-step pruning}
   \Statex {\bf Output:} Final vocabulary $\vocab$ and probabilities $p(\token_i)$.
\algsubsection{Conceptual algorithm from \citet{kudo-2018-subword}} 
\Statex {\bf Initialize:} Heuristically make a reasonably big seed vocabulary from the training corpus
\While{$|\vocab| > \vocabsize$} \Comment{Pruning Loop}
  \State (a) Fixing the set of vocabulary, optimize $p(\token)$ with the EM algorithm.
  \State (b) Compute the loss for each subword $x_i$, which represents 
        how likely the likelihood $\loss$ is reduced when the subword $x_i$ is 
        removed from the current vocabulary.
  \State (c) Sort the symbols by loss, and keep the top 80\% of subwords.   
  \EndWhile
\algsubsection{SentencePiece Implementation}    
   \State {\bf Initialize:} $\vocab \leftarrow \vocabseed(\mandatory{\corpus}, \configurable{\vocabsizeseed})$ using suffix arrays (Algorithm~\ref{alg:init}).
   \State $\ninterim \leftarrow \hardcoded{\overshootfactor} \cdot n$ \Comment{Set an intermediate target size slightly larger than final.}
   \While{True} \Comment{Pruning loop}
   \For{$i = 1$ \textbf{to} $\nem$} \Comment{EM sub-iterations}
   \State {\bf E-step:} Compute expected counts $c_j$ for each token $\token_j \in \vocab$ using forward-backward algorithm
   \State {\bf Fast-prune step:} Remove tokens $\token_j$ with expected count $c_j < \hardcoded{\mprunecount}$
   \State {\bf M-step:} Update token probabilities $p(\token_j) \leftarrow \digamma(c_j) / \sum_k \digamma(c_k)$ \Comment{digamma function $\digamma$}
   \EndFor
   \State \IIf{$|\vocab| \leq \ninterim$}{~\algorithmicbreak}
   \State {\bf Pruning step:}
   \State \quad $n_{\text{new}} \leftarrow \max(\ninterim,~\configurable{\alphaprune} \cdot |\vocab| )$
   \State \quad For each token $\token_j$, calculate the effect on $\loss$ if it were replaced by its optimal sub-tokenization
   \State \quad Prune $\vocab$ to size $n_{\text{new}}$ by removing tokens with the lowest effect on the objective $\loss$ if removed
   \EndWhile
   \State {\bf Finalize:} Flat Score Pruning of vocabulary $V$ to size $\vocabsize$ by selecting those with highest $p(\token_i)$
\end{algorithmic}

%% file: algorithm_app_init.tex
\section{Seed Vocabulary Creation Algorithms\label{app:init}}

The algorithm below uses a suffix array to find frequent substrings efficiently.
A \emph{suffix array} sorts all suffixes of the corpus lexicographically, placing suffixes with shared prefixes adjacent to each other.
The \emph{longest common prefix} (LCP) between adjacent suffixes tells us how many characters they share.
The algorithm maintains a stack to track ranges of suffixes sharing a common prefix, known as \say{open intervals}.
When the LCP drops below a stack entry's height, that prefix's interval has closed, and we emit the pattern with its frequency.

\begin{algorithm}
\caption{SentencePiece Seed Vocabulary Creation}
\label{alg:init}
\begin{algorithmic}[1]
 \State {\bf Input:} Corpus $\corpus$, maximum token length $\maxlength$, Target seed vocabulary size $\vocabsizeseed$
 \State {\bf Output:} Seed vocabulary $\vocabseed$ with initial scores
 \State \textbf{Initialization}
 \State $\corpusconcat \leftarrow$ Concatenation of $\corpus$, separated by boundary markers.
 \State SA $\leftarrow \operatorname{SuffixArray}(\corpusconcat)$ \Comment{SA[r] is the index of r\textsuperscript{th} suffix in lexicographical order} 
 \State $\vocabseed \leftarrow \text{empty queue}$ \Comment{Bounded priority queue, size $\vocabsizeseed$, sorted by score}
 \State $\stack \leftarrow \text{empty stack}$
 \State \textbf{Stack-based prefix emission loop}
 \For{$\rcurr = 1$ to $|\suffixarray| + 1$} \Comment{Process all adjacent suffix pairs in lexicographical order}
    \State $h \leftarrow \operatorname{LCP}(\suffixarray[\rcurr-1], \suffixarray[\rcurr])$ \Comment{Length of common prefix. $h=0$ in last iteration}
    \State $\rleftboundstack \leftarrow \rcurr - 1$
    \While{$\algorithmicnot{} \operatorname{IsEmpty}(\stack)~\algorithmicand{} \operatorname{Top}(\stack).height > h$}\label{line:stackloopstart}
        \State $(\ell, \rstart) \leftarrow \operatorname{Pop}(\stack)$ \Comment{pattern length $\ell$, interval start}\label{line:pop}
        \State $\text{freq} \leftarrow \rcurr - \rstart$ \Comment{Number of suffixes sharing prefix of length $\geq \ell$}
        \State $p \leftarrow$ prefix of length $\ell$ of suffix at $\suffixarray[\rstart]$
        \If{$1 < \ell \leq \maxlength$ \algorithmicand{} $\text{freq} \geq 2$ \algorithmicand{} $p$ is a valid token}
            \State Add $\left(p, \text{freq} \times \ell\right)$ to $\vocabseed$ \Comment{Character coverage score: frequency $\times$ length}
        \EndIf\label{line:add}
        \State $\rleftboundstack \leftarrow  \rstart$ \Comment{Propagate interval start for merging}
    \EndWhile\label{line:stackloopend}
    
    \If{IsEmpty(\stack) \algorithmicor{} Top(\stack).height $< h$}
        \State Push $\left(\text{height}=h, \text{rank=}\rleftboundstack\right)$ onto $\stack$
    \EndIf
 \EndFor
 \State \textbf{return} $\vocabseed$
\end{algorithmic}
\end{algorithm}

\begin{algorithm}
\caption{Fallback variant: including longest valid prefix of rejected tokens when creating seed vocabulary. (replacing lines \ref{line:pop}--\ref{line:add} in \autoref{alg:init})}
\label{alg:init:repair}
\begin{algorithmic}[1]
\ContinuedFloat
    \State $(\ell, \rstart) \leftarrow \operatorname{Pop}(\stack)$ \Comment{Pattern length $\ell$, interval start rank}
    \State $\text{freq} \leftarrow \rcurr - \rstart$
    \State $\ell_s \leftarrow \operatorname{Top}(\stack).\text{height} + 1$ \Comment{if stack is empty, use $h+1$}
    \For{$\ell_i = \ell$ down to $\ell_s$} \Comment{\textbf{Change:} Iterate over all intermediate lengths}
        \State $p \leftarrow$ prefix of length $\ell_i$ from suffix at $\suffixarray[\rstart]$
        \If{$1 < \ell_i \leq \maxlength$ \algorithmicand{} $\text{freq} \geq 2$ \algorithmicand{} $p$ is a valid token}
            \State $\text{score} \leftarrow \text{freq} \times \ell_i$
            \State Add $(p, \text{score})$ to $\vocabseed$
            \State \textbf{break $\ell_i$ loop} \Comment{Only include the longest valid token}
        \EndIf
    \EndFor
\end{algorithmic}
\end{algorithm}

%% file: tables_app_hyper.tex
\section{Detailed Results of Unigram Hyperparameter Variation\label{app:hyperparams}}

\autoref{tab:app-seed} presents results for the seed vocabulary algorithm; \autoref{tab:pruning-loop} for pruning loop parameters; \autoref{tab:res-fsp} for \FSP{}; and \autoref{tab:unigram_hyperparams} for other Unigram hyperparameters.
Loss and token count are presented as relative change vs. baseline, while vocabulary overlap as absolute \%, where 100\% = identical to baseline.
\goodoutlier{Good} and \badoutlier{bad} changes above 0.5\% are highlighted along with \biggestoutlier{the largest} absolute change per parameter.


\begin{table*}[hb]
\centering
\small
\begin{tabular}{ll|rrr|rrr}
\toprule
 &  & \multicolumn{3}{c|}{30\,MB} & \multicolumn{3}{c}{300\,MB} \\
\cmidrule(lr){3-5} \cmidrule(lr){6-8}
  &   & Full text & Full text & Pretokens & Full text & Full text & Pretokens \\
Corpus & Metric &  & recovery & recovery &  & recovery & recovery \\
\midrule
\multirow{3}{*}{English} & \textit{Loss} & +0.27 & +0.04 & -0.01 & -0.00 & +0.02 & +0.01 \\
  & \textit{Tokens} & +0.26 & +0.01 & -0.00 & -0.01 & +0.01 & +0.01 \\
  & \textit{Vocabulary Overlap} & 82.3 & 89.6 & 93.9 & 96.5 & 91.5 & 91.5 \\
\cmidrule{2-8}
\multirow{3}{*}{German} & \textit{Loss} & +0.14 & +0.17 & +0.21 & +0.01 & +0.17 & \biggestoutlier{+0.19} \\
  & \textit{Tokens} & +0.32 & +0.02 & +0.04 & -0.01 & -0.03 & -0.02 \\
  & \textit{Vocabulary Overlap} & 89.4 & 80.3 & 79.9 & 98.6 & 82.3 & \biggestoutlier{81.3} \\
\cmidrule{2-8}
\multirow{3}{*}{Arabic} & \textit{Loss} & +0.39 & +0.02 & +0.00 & +0.02 & +0.00 & -0.00 \\
  & \textit{Tokens} & \badoutlier{+1.24} & -0.05 & -0.09 & +0.03 & -0.01 & +0.01 \\
  & \textit{Vocabulary Overlap} & 74.4 & 83.5 & 84.8 & 95.6 & 89.7 & 89.6 \\
\cmidrule{2-8}
\multirow{3}{*}{Hindi} & \textit{Loss} & \biggestoutlier{\badoutlier{+2.10}} & -0.02 & -0.03 & +0.09 & -0.00 & -0.00 \\
  & \textit{Tokens} & \biggestoutlier{\badoutlier{+2.29}} & -0.01 & -0.01 & \biggestoutlier{+0.23} & -0.00 & +0.00 \\
  & \textit{Vocabulary Overlap} & \biggestoutlier{57.9} & 88.1 & 89.0 & 89.5 & 90.9 & 91.2 \\
\cmidrule{2-8}
\multirow{3}{*}{Korean} & \textit{Loss} & +0.33 & +0.01 & +0.01 & +0.02 & +0.02 & +0.01 \\
  & \textit{Tokens} & \badoutlier{+1.24} & -0.06 & -0.03 & +0.09 & -0.01 & -0.00 \\
  & \textit{Vocabulary Overlap} & 82.1 & 93.0 & 95.8 & 96.1 & 96.4 & 97.0 \\
\cmidrule{2-8}
\multirow{3}{*}{Chinese} & \textit{Loss} & +0.39 & +0.02 & +0.01 & +0.09 & +0.01 & +0.00 \\
  & \textit{Tokens} & +0.29 & +0.02 & +0.01 & +0.07 & -0.00 & +0.00 \\
  & \textit{Vocabulary Overlap} & 88.6 & 97.4 & 98.0 & 97.7 & 98.9 & 99.3 \\
\cmidrule{2-8}
\multirow{3}{*}{Mean} & \textit{Loss} & \badoutlier{+0.60} & +0.04 & +0.03 & +0.04 & +0.04 & +0.04 \\
  & \textit{Tokens} & \badoutlier{+0.94} & -0.01 & -0.02 & +0.07 & -0.01 & +0.00 \\
  & \textit{Vocabulary Overlap} & 79.1 & 88.6 & 90.2 & 95.7 & 91.6 & 91.6 \\
\bottomrule
\end{tabular}
\caption{\label{tab:app-seed}Seed vocabulary algorithm comparison: relative performance compared to pretoken-based.}
\end{table*}

\begin{table*}[hb]
\centering
\small
\begin{tabular}{l|r|rrr|rrr|rrr}
\toprule
     & \multicolumn{1}{c}{digamma (on)} & \multicolumn{3}{c}{$\mprunecount$ (0.5)} & \multicolumn{3}{c}{$\nem$ (2)} & \multicolumn{3}{c}{$\alphaprune$ (0.75)} \\
\cmidrule(lr){2-2} \cmidrule(lr){3-5} \cmidrule(lr){6-8} \cmidrule(lr){9-11}
    Metric & Off & 0.0 & 2.0 & 10.0 & 1 & 3 & 5 & 0.5 & 0.9 & 0.95 \\
\midrule
  \textit{Loss} & -0.01 & +0.00 & +0.02 & -0.00 & +0.02 & -0.00 & -0.00 & +0.15 & -0.07 & -0.09 \\
  \textit{Tokens} & -0.00 & +0.00 & +0.00 & -0.03 & +0.02 & -0.00 & +0.00 & -0.04 & +0.03 & +0.03 \\
  \textit{Vocab. Overlap} & 98.0 & 94.0 & 93.8 & 93.7 & 93.3 & 95.7 & 94.1 & 88.8 & 93.2 & 92.4 \\
\bottomrule
\end{tabular}
\caption{\label{tab:pruning-loop}Pruning loop parameters: M-step digamma transformation (default on), pruning threshold (default $\mprunecount=0.5$), EM sub-iterations (default $\nem=2$), and pruning shrinking factor (default $\alphaprune=0.75$). Mean results over six monolingual corpora.}
\end{table*}

\begin{table*}
\centering
\small
\begin{tabular}{ll|rrrrrr}
\toprule
Corpus & Metric & 0.0 & 0.25 & 0.5 & 0.75 & 0.9 & 0.95 \\
\midrule
\multirow{3}{*}{English} & \textit{Loss} & \badoutlier{+0.72} & \badoutlier{+0.64} & \badoutlier{+0.59} & \badoutlier{+0.55} & \badoutlier{+0.53} & \badoutlier{+0.53} \\
  & \textit{Tokens} & +0.34 & -0.06 & -0.21 & -0.24 & -0.25 & -0.25 \\
  & \textit{Vocabulary Overlap} & 73.4 & 75.8 & 75.9 & 75.5 & 74.5 & 74.4 \\
\cmidrule{2-8}
\multirow{3}{*}{German} & \textit{Loss} & \badoutlier{+1.41} & \badoutlier{+1.06} & \badoutlier{+0.96} & \badoutlier{+0.85} & \badoutlier{+0.79} & \badoutlier{+0.77} \\
  & \textit{Tokens} & +0.19 & -0.47 & \goodoutlier{-0.60} & \goodoutlier{-0.70} & \goodoutlier{-0.72} & \goodoutlier{-0.73} \\
  & \textit{Vocabulary Overlap} & 69.0 & 70.4 & 70.9 & 70.5 & 69.9 & 69.7 \\
\cmidrule{2-8}
\multirow{3}{*}{Arabic} & \textit{Loss} & \badoutlier{+1.19} & \badoutlier{+0.82} & \badoutlier{+0.74} & \badoutlier{+0.57} & +0.49 & +0.47 \\
  & \textit{Tokens} & +0.19 & \goodoutlier{-0.97} & \goodoutlier{-1.02} & \goodoutlier{-0.93} & \goodoutlier{-0.85} & \goodoutlier{-0.84} \\
  & \textit{Vocabulary Overlap} & 65.1 & 71.3 & 72.9 & 74.7 & 74.9 & 74.8 \\
\cmidrule{2-8}
\multirow{3}{*}{Hindi} & \textit{Loss} & \badoutlier{+0.66} & +0.49 & +0.47 & +0.45 & +0.44 & +0.43 \\
  & \textit{Tokens} & \badoutlier{+0.53} & -0.00 & -0.11 & -0.13 & -0.13 & -0.14 \\
  & \textit{Vocabulary Overlap} & 74.5 & 77.2 & 77.6 & 77.0 & 76.3 & 76.2 \\
\cmidrule{2-8}
\multirow{3}{*}{Korean} & \textit{Loss} & \biggestoutlier{\badoutlier{+1.77}} & \badoutlier{+1.08} & \badoutlier{+0.95} & \badoutlier{+0.84} & \badoutlier{+0.79} & \badoutlier{+0.77} \\
  & \textit{Tokens} & -0.10 & \goodoutlier{-1.14} & \goodoutlier{-1.23} & \goodoutlier{-1.30} & \goodoutlier{-1.31} & \biggestoutlier{\goodoutlier{-1.32}} \\
  & \textit{Vocabulary Overlap} & \biggestoutlier{65.0} & 70.1 & 70.8 & 71.4 & 71.5 & 71.5 \\
\cmidrule{2-8}
\multirow{3}{*}{Chinese} & \textit{Loss} & \badoutlier{+1.05} & \badoutlier{+1.00} & \badoutlier{+0.97} & \badoutlier{+0.94} & \badoutlier{+0.93} & \badoutlier{+0.93} \\
  & \textit{Tokens} & -0.09 & -0.25 & -0.26 & -0.31 & -0.33 & -0.33 \\
  & \textit{Vocabulary Overlap} & 74.6 & 74.2 & 74.4 & 74.1 & 74.0 & 73.8 \\
\cmidrule{2-8}
\multirow{3}{*}{Mean} & \textit{Loss} & \badoutlier{+1.13} & \badoutlier{+0.85} & \badoutlier{+0.78} & \badoutlier{+0.70} & \badoutlier{+0.66} & \badoutlier{+0.65} \\
  & \textit{Tokens} & +0.18 & -0.48 & \goodoutlier{-0.57} & \goodoutlier{-0.60} & \goodoutlier{-0.60} & \goodoutlier{-0.60} \\
  & \textit{Vocabulary Overlap} & 70.3 & 73.2 & 73.8 & 73.9 & 73.5 & 73.4 \\
\bottomrule
\end{tabular}
\caption{\label{tab:res-fsp}\FSP{}: relative performance compared to default settings, using varying shrinking factor $\alphaprune$. We use $\overshootfactor=1$ for these experiments,
as it would be identical to normal pruning steps.}
\end{table*}

\begin{table*}
\centering\small
\begin{tabular}{ll|rrrr|rrrr}
\toprule
  &   & \multicolumn{4}{c}{$\vocabsizeseedfac$ (10)} & \multicolumn{4}{c}{$\overshootfactor$ (1.1)} \\
\cmidrule(lr){3-6} \cmidrule(lr){7-10}
Corpus & Metric & 3 & 25 & 50 & 100 & 1.0 & 1.25 & 1.5 & 2.0 \\
\midrule
\multirow{3}{*}{Arabic} & \textit{Loss} & \badoutlier{+0.60} & +0.03 & +0.02 & +0.02 & -0.30 & +0.44 & \badoutlier{+1.00} & \badoutlier{+2.00} \\
  & \textit{Tokens} & +0.46 & -0.17 & -0.20 & -0.20 & +0.26 & -0.21 & -0.37 & \goodoutlier{-0.50} \\
  & \textit{Vocabulary Overlap} & 67.1 & 86.7 & 87.3 & 87.3 & 88.0 & 89.0 & 83.0 & 78.2 \\
\cmidrule{2-10}
\multirow{3}{*}{German} & \textit{Loss} & \biggestoutlier{\badoutlier{+2.90}} & -0.09 & -0.10 & -0.11 & -0.23 & +0.45 & \badoutlier{+1.10} & \badoutlier{+2.09} \\
  & \textit{Tokens} & \biggestoutlier{\badoutlier{+2.51}} & -0.06 & -0.09 & -0.09 & +0.21 & -0.14 & -0.33 & -0.35 \\
  & \textit{Vocabulary Overlap} & \biggestoutlier{45.7} & 87.1 & 86.4 & 86.6 & 87.1 & 88.2 & 78.8 & \biggestoutlier{73.6} \\
\cmidrule{2-10}
\multirow{3}{*}{English} & \textit{Loss} & \badoutlier{+0.92} & -0.02 & -0.03 & -0.03 & -0.21 & +0.32 & \badoutlier{+0.71} & \badoutlier{+1.21} \\
  & \textit{Tokens} & \badoutlier{+0.59} & -0.01 & -0.02 & -0.02 & +0.07 & -0.04 & -0.10 & -0.11 \\
  & \textit{Vocabulary Overlap} & 62.1 & 92.2 & 92.4 & 92.4 & 87.4 & 89.3 & 83.0 & 79.2 \\
\cmidrule{2-10}
\multirow{3}{*}{Hindi} & \textit{Loss} & \badoutlier{+0.79} & -0.01 & -0.01 & -0.01 & -0.20 & +0.31 & \badoutlier{+0.63} & \badoutlier{+1.07} \\
  & \textit{Tokens} & \badoutlier{+0.57} & -0.01 & -0.01 & -0.01 & +0.05 & -0.00 & -0.05 & -0.07 \\
  & \textit{Vocabulary Overlap} & 65.2 & 91.4 & 91.4 & 91.4 & 87.5 & 89.9 & 84.4 & 80.9 \\
\cmidrule{2-10}
\multirow{3}{*}{Korean} & \textit{Loss} & \badoutlier{+0.63} & -0.01 & +0.01 & +0.03 & -0.38 & \badoutlier{+0.52} & \badoutlier{+1.14} & \biggestoutlier{\badoutlier{+2.14}} \\
  & \textit{Tokens} & \goodoutlier{-0.59} & +0.09 & +0.14 & +0.14 & +0.26 & -0.31 & \goodoutlier{-0.60} & \biggestoutlier{\goodoutlier{-0.74}} \\
  & \textit{Vocabulary Overlap} & 77.1 & 91.8 & 91.5 & 91.8 & 87.0 & 87.6 & 79.2 & 74.0 \\
\cmidrule{2-10}
\multirow{3}{*}{Chinese} & \textit{Loss} & \badoutlier{+0.80} & -0.04 & -0.04 & -0.03 & -0.47 & \badoutlier{+0.66} & \badoutlier{+1.29} & \badoutlier{+2.06} \\
  & \textit{Tokens} & -0.13 & +0.03 & +0.04 & +0.04 & +0.02 & +0.10 & +0.03 & -0.08 \\
  & \textit{Vocabulary Overlap} & 75.7 & 96.4 & 96.5 & 96.4 & 87.0 & 88.0 & 81.0 & 76.7 \\
\cmidrule{1-10}
\multirow{3}{*}{Mean} & \textit{Loss} & \badoutlier{+1.11} & -0.02 & -0.03 & -0.02 & -0.30 & +0.45 & \badoutlier{+0.98} & \badoutlier{+1.76} \\
  & \textit{Tokens} & \badoutlier{+0.57} & -0.02 & -0.02 & -0.02 & +0.15 & -0.10 & -0.24 & -0.31 \\
  & \textit{Vocabulary Overlap} & 65.5 & 90.9 & 90.9 & 91.0 & 87.4 & 88.7 & 81.6 & 77.1 \\
\bottomrule
\end{tabular}
\caption{\label{tab:unigram_hyperparams} Hyperparameter results: relative seed size (default $\vocabsizeseedfac=10$), pruning overshoot (default $\overshootfactor=1.1$).}
\end{table*}

%% file: table_app_morphscore.tex
\section{MorphScore Evaluation}

\label{app:morphscore}

\autoref{tab:morphscore-examples} shows MorphScore \citep[version 2,][]{arnett2025evaluatingmorphologicalalignmenttokenizers} evaluation results comparing our Baseline unigram tokenizer, the FSP (\FSP{}) variant, and BPE on English morphological segmentation.

\begin{table}[H]
\centering
\small
\begin{tabular}{@{}lrp{2.0cm}p{2.0cm}p{2.0cm}p{2.0cm}@{}}
\toprule
\textbf{Word} & \textbf{Freq} & \textbf{Gold} & \textbf{Baseline} & \textbf{FSP} & \textbf{BPE} \\
\midrule
\multicolumn{6}{l}{\textbf{All Methods Match Gold} (258 words covering 9.5\% of corpus)} \\
\midrule
been & 2.2\% & be | en & \matchcolor{be | en} & \matchcolor{be | en} & \matchcolor{be | en} \\
months & 0.4\% & month | s & \matchcolor{month | s} & \matchcolor{month | s} & \matchcolor{month | s} \\
following & 0.3\% & follow | ing & \matchcolor{follow | ing} & \matchcolor{follow | ing} & \matchcolor{follow | ing} \\
\midrule
\multicolumn{6}{l}{\textbf{Only Baseline Matches} (443 words covering 6.8\% of corpus)} \\
\midrule
attached & 0.4\% & attach | ed & \matchcolor{attach | ed} & atta | ched & att | ached \\
Minutes & 0.2\% & Minute | s & \matchcolor{Min | ute | s} & Min | utes & Min | utes \\
animals & 0.2\% & animal | s & \matchcolor{animal | s} & anim | als & anim | als \\
\midrule
\multicolumn{6}{l}{\textbf{Only FSP Matches} (111 words covering 1.3\% of corpus)} \\
\midrule
Palestinians & 0.1\% & Palestinian | s & P | ale | stin | ians & \matchcolor{Pal | est | inian | s} & Pal | est | in | ians \\
instructions & 0.1\% & instruction | s & in | struct | ions & \matchcolor{in | struction | s} & inst | ru | ctions \\
quoted & 0.1\% & quote | d & quot | ed & \matchcolor{quote | d} & qu | oted \\
\midrule
\multicolumn{6}{l}{\textbf{Only BPE Matches} (68 words covering 0.8\% of corpus)} \\
\midrule
intended & 0.1\% & intend | ed & int | ended & int | ended & \matchcolor{intend | ed} \\
bigger & 0.1\% & big | ger & b | igger & b | igger & \matchcolor{big | ger} \\
seemed & 0.1\% & seem | ed & see | med & see | med & \matchcolor{se | em | ed} \\
\midrule
\multicolumn{6}{l}{\textbf{No Method Matches} (1000 words covering 12.6\% of corpus)} \\
\midrule
comments & 0.2\% & comment | s & \nomatchcolor{com | ments} & \nomatchcolor{com | ments} & \nomatchcolor{com | ments} \\
weapons & 0.2\% & weapon | s & \nomatchcolor{we | ap | ons} & \nomatchcolor{we | ap | ons} & \nomatchcolor{we | ap | ons} \\
decided & 0.1\% & decide | d & \nomatchcolor{de | cid | ed} & \nomatchcolor{de | cid | ed} & \nomatchcolor{dec | ided} \\
\bottomrule
\end{tabular}
\caption{MorphScore examples for English, grouped by which methods match gold morpheme boundaries. Each category shows the three most frequent words meeting that criterion. Freq is the word's share of total evaluation frequency.}
\label{tab:morphscore-examples}
\end{table}

%% file: references.bib
@misc{penedo2025finewiki,
  author    = {Guilherme Penedo},
  title     = {{FineWiki}},
  year      = {2025},
  publisher = {Hugging Face Datasets},
  url       = {https://huggingface.co/datasets/HuggingFaceFW/finewiki},
  urldate   = {2025-10-20},
  note      = {Source: Wikimedia Enterprise Snapshot API (https://api.enterprise.wikimedia.com/v2/snapshots). Text licensed under CC BY-SA 4.0 with attribution to Wikipedia contributors.}
}

@misc{provilkov2020bpedropoutsimpleeffectivesubword,
      title={{BPE-Dropout}: Simple and Effective Subword Regularization}, 
      author={Ivan Provilkov and Dmitrii Emelianenko and Elena Voita},
      year={2020},
      eprint={1910.13267},
      archivePrefix={arXiv},
      primaryClass={cs.CL},
      url={https://arxiv.org/abs/1910.13267}, 
}

@misc{arnett2025evaluatingmorphologicalalignmenttokenizers,
      title={Evaluating Morphological Alignment of Tokenizers in 70 Languages}, 
      author={Catherine Arnett and Marisa Hudspeth and Brendan O'Connor},
      year={2025},
      eprint={2507.06378},
      archivePrefix={arXiv},
      primaryClass={cs.CL},
      url={https://arxiv.org/abs/2507.06378}, 
}

@unpublished{clarablog,
    author = {Clara Meister},
    title = {{UnigramLM}: An Attempt at Writing The Missing Manual},
    year = {2025},
    note = {Blog post},
    url = {https://cimeister.github.io/blog/unigramlm/}
}

@inproceedings{scriptbpe,
      title={{BPE} Stays on {SCRIPT}: Structured Encoding for Robust Multilingual Pretokenization}, 
      author={Sander Land and Catherine Arnett},
      year={2025},
      booktitle = {ICML 2025 Tokenization Workshop},
      url={https://openreview.net/forum?id=AO78CqwaUO}
}

@misc{chizhov2024bpegetspickyefficient,
      title={{BPE} Gets Picky: Efficient Vocabulary Refinement During Tokenizer Training}, 
      author={Pavel Chizhov and Catherine Arnett and Elizaveta Korotkova and Ivan P. Yamshchikov},
      year={2024},
      eprint={2409.04599},
      archivePrefix={arXiv},
      primaryClass={cs.CL},
      url={https://arxiv.org/abs/2409.04599}, 
}

@inproceedings{schmidt-etal-2024-tokenization,
    title = "Tokenization Is More Than Compression",
    author = "Schmidt, Craig W  and
      Reddy, Varshini  and
      Zhang, Haoran  and
      Alameddine, Alec  and
      Uzan, Omri  and
      Pinter, Yuval  and
      Tanner, Chris",
    editor = "Al-Onaizan, Yaser  and
      Bansal, Mohit  and
      Chen, Yun-Nung",
    booktitle = "Proceedings of the 2024 Conference on Empirical Methods in Natural Language Processing",
    month = nov,
    year = "2024",
    address = "Miami, Florida, USA",
    publisher = "Association for Computational Linguistics",
    url = "https://aclanthology.org/2024.emnlp-main.40/",
    doi = "10.18653/v1/2024.emnlp-main.40",
    pages = "678--702",
}

@inproceedings{sennrich-etal-2016-neural,
    title = "Neural Machine Translation of Rare Words with Subword Units",
    author = "Sennrich, Rico  and
      Haddow, Barry  and
      Birch, Alexandra",
    editor = "Erk, Katrin  and
      Smith, Noah A.",
    booktitle = "Proceedings of the 54th Annual Meeting of the Association for Computational Linguistics (Volume 1: Long Papers)",
    month = aug,
    year = "2016",
    address = "Berlin, Germany",
    publisher = "Association for Computational Linguistics",
    url = "https://aclanthology.org/P16-1162/",
    doi = "10.18653/v1/P16-1162",
    pages = "1715--1725"
}

@inproceedings{kudo-richardson-2018-sentencepiece,
    title = "{SentencePiece}: A simple and language independent subword tokenizer and detokenizer for Neural Text Processing",
    author = "Kudo, Taku  and
      Richardson, John",
    editor = "Blanco, Eduardo  and
      Lu, Wei",
    booktitle = "Proceedings of the 2018 Conference on Empirical Methods in Natural Language Processing: System Demonstrations",
    month = nov,
    year = "2018",
    address = "Brussels, Belgium",
    publisher = "Association for Computational Linguistics",
    url = "https://aclanthology.org/D18-2012/",
    doi = "10.18653/v1/D18-2012",
    pages = "66--71",
}

@inproceedings{kudo-2018-subword,
    title = "Subword Regularization: Improving Neural Network Translation Models with Multiple Subword Candidates",
    author = "Kudo, Taku",
    editor = "Gurevych, Iryna  and
      Miyao, Yusuke",
    booktitle = "Proceedings of the 56th Annual Meeting of the Association for Computational Linguistics (Volume 1: Long Papers)",
    month = jul,
    year = "2018",
    address = "Melbourne, Australia",
    publisher = "Association for Computational Linguistics",
    url = "https://aclanthology.org/P18-1007",
    doi = "10.18653/v1/P18-1007",
    pages = "66--75",
}

@article{baum1972inequality,
  title={An inequality and associated maximization technique in statistical estimation for probabilistic functions of Markov processes},
  author={Baum, Leonard E},
  journal={Inequalities},
  volume={3},
  number={1},
  pages={1--8},
  year={1972}
}
